\documentclass{article}

\usepackage{arxiv}

\usepackage[utf8]{inputenc} % allow utf-8 input
\usepackage[T1]{fontenc}    % use 8-bit T1 fonts
\usepackage{hyperref}       % hyperlinks
\usepackage{url}            % simple URL typesetting
\usepackage{booktabs}       % professional-quality tables
\usepackage{amsfonts}       % blackboard math symbols
\usepackage{nicefrac}       % compact symbols for 1/2, etc.
\usepackage{microtype}      % microtypography
\usepackage{lipsum}		% Can be removed after putting your text content
\usepackage{graphicx}
\usepackage{natbib}
\usepackage{doi}
\usepackage[linesnumbered,ruled,vlined]{algorithm2e}
\usepackage{multirow}

\title{Neuro-symbolic Zero-Shot Code Cloning with Cross-Language Intermediate Representation}

\author{ % \href{https://orcid.org/0000-0000-0000-0000}
{Krishnam Hasija}%\thanks{Use footnote for providing further
        \\
	TCS Research\\
	\texttt{krishnam.hasija@tcs.com} \\
	\And
 {Shrishti Pradhan} \\
	TCS Research\\
	\texttt{shrishti.pradhan@tcs.com} \\
 \And
 {Manasi Patwardhan} \\
	TCS Research\\
	\texttt{manasi.patwardhan@tcs.com} \\
 \And
  {Raveendra Kumar Medicherla} \\
	TCS Research\\
	\texttt{raveendra.kumar@tcs.com} \\ 
  \And
 {Lovekesh Vig} \\
	TCS Research\\
	\texttt{lovekesh.vig@tcs.com} \\
  \And
  {Ravindra Naik} \\
	TCS Research\\
	\texttt{rd.naik@tcs.com} \\
}

\renewcommand{\shorttitle}{\textit{arXiv} Template}

\hypersetup{
pdftitle={A template for the arxiv style},
pdfsubject={q-bio.NC, q-bio.QM},
pdfauthor={David S.~Hippocampus, Elias D.~Striatum},
pdfkeywords={First keyword, Second keyword, More},
}

\begin{document}
\maketitle

\begin{abstract}
	%\lipsum[1]
In this paper, we define a neuro-symbolic approach to address the task of finding semantically similar clones for the codes of the legacy programming language COBOL, without
%no availability of 
training data. We define a meta-model that is instantiated to have an Intermediate Representation (IR) in the form of Abstract Syntax Trees (ASTs) common across codes in C and COBOL. We linearize the IRs using Structure Based Traversal (SBT) to create sequential inputs. We further fine-tune UnixCoder, the best-performing model for zero-shot cross-programming language code search, for the Code Cloning task with the SBT IRs of C code-pairs, available in the CodeNet dataset. This allows us to learn latent representations for the IRs of the C codes, which are transferable to the IRs of the COBOL codes. With this fine-tuned UnixCoder, we get a performance improvement of 12.85 MAP@2 over the pre-trained UniXCoder model, in a zero-shot setting, on the COBOL test split synthesized from the CodeNet dataset. This demonstrates the efficacy of our meta-model based approach to facilitate cross-programming language transfer.   
\end{abstract}

% keywords can be removed

\section{Introduction}
%\lipsum[2]
%\lipsum[3]
Recent advancements in pre-training Language Models (LMs) for learning code representations 
%\cite{Lu2021CodeXGLUEAM,Jiang2021TreeBERTAT,Wang2021CodeT5IU,Guo2022UniXcoderUC,Wang2022CODEMVPLT,Wang2021SynCoBERTSM, Wang2022GypSumLH} % CodeGPT, TreeBERT, CodeT5, UnixCoder, Code MVP, SYNCOBERT , GypSum 
have lead to improvements in the down-stream code understanding and generation tasks \cite{Ahmad2021UnifiedPF,Feng2020CodeBERTAP,Guo2021GraphCodeBERTPC,Lu2021CodeXGLUEAM,Wang2021SynCoBERTSM,Guo2022UniXcoderUC,Wang2022CODEMVPLT,Wang2021CodeT5IU,Qi2021ProphetNetXLP,Jiang2021TreeBERTAT,Wang2022GypSumLH}. % such as Code Search \cite{Feng2020CodeBERTAP, Guo2021GraphCodeBERTPC, Lu2021CodeXGLUEAM, Wang2021SynCoBERTSM, Guo2022UniXcoderUC, Wang2022CODEMVPLT}, Code clone detection \cite{Guo2021GraphCodeBERTPC, Ahmad2021UnifiedPF, Lu2021CodeXGLUEAM, Wang2021CodeT5IU, Wang2021SynCoBERTSM, Guo2022UniXcoderUC, Wang2022CODEMVPLT}, Code Summarization  \cite{Feng2020CodeBERTAP,Ahmad2021UnifiedPF,Qi2021ProphetNetXLP, Lu2021CodeXGLUEAM, Jiang2021TreeBERTAT, Wang2021CodeT5IU, Guo2022UniXcoderUC, Wang2022GypSumLH}, Code Generation \cite{Ahmad2021UnifiedPF, Lu2021CodeXGLUEAM, Wang2021CodeT5IU, Guo2022UniXcoderUC}, Code Completion \cite{Svyatkovskiy2020IntelliCodeCC, Lu2021CodeXGLUEAM, Guo2022UniXcoderUC}, Code Refinement \cite{Guo2021GraphCodeBERTPC, Buratti2020ExploringSN, Wang2021CodeT5IU} and Code Translation \cite{Guo2021GraphCodeBERTPC, Lu2021CodeXGLUEAM, Wang2021CodeT5IU, Wang2021SynCoBERTSM}. 
These LMs are trained with large volumes of monolingual code data (in the order of hundreds of GBs) with pre-training objectives such as Masked Language Modeling (MLM), Replaced Token Detection (RTD), Denoising objectives (DNS), etc. %Unidirectional Language Modelling (ULM), Denoising objectives (DNS), Multi-modal Contrastive Learning (MCL), Cross-Modal Generation (CMG), Masked Span Prediction (MSP), Masked Identifier Prediction (MIP), dataflow edge prediction or type inference, and multi-view contrastive pre-training. %The models require lot of pre-training data in the order of hundreds of GBs. For the legacy languages like COBOL, to the best of our knowledge, there are no publicly available datasets except CodeNet \cite{Puri2021ProjectCA}, which has COBOL data. Moreover, CodeNet has only 727 accepted COBOL submissions for 325 problem descriptions. This makes it hard to pre-train language models for legacy languages like COBOL. 
%More recently, Large Language Models (LLMs) such as T5\cite{48643}, GPT3\cite{brown2020language}, Codex\cite{Chen2021EvaluatingLL}, PaLM\cite{chowdhery2022palm}, pretrained with massive volumes (in the order on 50TB) of data have shown improved performance for code understanding tasks using in-context learning \cite{Brown2020LanguageMA,Huang2022TowardsRI}, which work in zero-shot or few-shot setting, with no or very less number of available samples for a given task. This can be a solution to perform various tasks for low-resource languages.  However, eventually these LLMs are not going to be available as cost-free solutions and also would require porting of the propriety code data to the server on which these LLMs reside, raising the recurrent cost and code privacy concerns. On the other hand, above listed smaller pretrained LMs are freely available and can reside on local server. However, in-context learning does not work well with these smaller LMs \cite{Qiu2022EvaluatingTI,Hosseini2022OnTC} and they require task specific fine-tuning to yield acceptable performance

The focus of this paper is on the task of Code Clone detection %or Code-to-Code retrieval 
specifically for the legacy low-resource Programming Language (PL) COBOL. %, for which very few samples of similar codes are present, which are insufficient for training language models. 
There is a very high volume of COBOL application code still in use by organizations and institutions worldwide. Research suggests that there are more than 800 billion lines of COBOL code currently in use and therefore it is crucial to maintain and enhance the COBOL code, until it is modernized using digital transformation. Code clone detection is used to measure the \emph{similarity} of code fragments and has direct applications in code reuse, code compaction by replacing code snippets with more compact code clones, copyright infringement detection etc. It aims to detect whether different code fragments have the same behavior (i.e. give similar outputs on similar inputs) irrespective of their surface form or structure and syntax. The task is typically achieved by learning the \emph{semantic representation} of the code. 
%plagiarism detection in code clone detection uses 
%by learning the semantic representations of the Code. Code clone detection is used to measure the similarity of code fragments and aims to detect whether they have the same semantics (i.e. give similar outputs on similar inputs). Code-to-Code search aims to retrieve a ranked list of semantically similar codes give a code snippet as the query. The surface form or the structure and the syntax of the queried code and the retrieved ones can be different. 

In the literature, there is work done on Code Clone detection \cite{Guo2021GraphCodeBERTPC, Ahmad2021UnifiedPF, Lu2021CodeXGLUEAM, Wang2021CodeT5IU, Wang2021SynCoBERTSM, Guo2022UniXcoderUC, Wang2022CODEMVPLT} and code search tasks \cite{Feng2020CodeBERTAP, Guo2021GraphCodeBERTPC, Lu2021CodeXGLUEAM, Wang2021SynCoBERTSM, Guo2022UniXcoderUC, Wang2022CODEMVPLT} for PLs such as Java, Javascript, Python, Ruby, Go, etc. by pre-training and task specific fine-tuning of language models. The pre-training data used here is typically the CodeSearchNet dataset \cite{Husain2019CodeSearchNetCE} consisting of 352 GB of Java and 224 GB of python codes. For code-clone detection task specific fine-tuning typically POJ-104 \cite{mou2016convolutional} and BigCloneBench \cite{svajlenko2014towards} datasets are used. POJ-104 dataset consists of 104 problems including 500 C/C++ codes for each problem. BigCloneBench dataset includes $\sim$900K training examples for Java programming language from 10 different functionalities. CSN dataset \cite{Husain2019CodeSearchNetCE}, which is typically used for fine-tuning for code search task has $\sim$906K training samples for Java, Javascript, PHP, Python, Go and Ruby PLs. Thus, to pre-train and fine-tune these language models, a huge amount of PL-specific data is used. On the other hand, for  legacy languages like COBOL the above non-COBOL datasets \cite{mou2016convolutional,svajlenko2014towards,Husain2019CodeSearchNetCE} are not useful and to the best of our knowledge, there are no publicly available datasets other than CodeNet with COBOL codes as part of the dataset. Even in the CodeNet dataset, there are only 727 COBOL codes for 325 problem descriptions. Codes belonging to the same problem descriptions can form clones of each other as they share semantics. However, this tiny COBOL data is insufficient for pre-training as well as fine-tuning LMs for COBOL code representation learning and for downstream tasks such as code cloning. 

In this paper, we use an Intermediate Representation (IR) synthesized using a pre-defined meta-model, which is common across C and COBOL PLs.  CodeNet dataset contains $\sim$300K accepted C code submissions for $\sim$3K problem descriptions. We perform the following transforms (i) To avoid any inductive biases created because of meaningful function and variable names and thus to help the model focus on the underlying logic of the code, we perform a semantics preserving transformation, by replacing function and variable names from the C codes with more generic non-meaningful words from the vocabulary such as $FUNC$ or $VAR$  (ii) Transform C codes to our pre-defined IR, which is a form of Abstract Syntax Tree (AST) and an instance of a predefined \emph{meta-model}, by a language-specific parser, (iii) Map specific code tokens in C which appear at the leaf nodes of the IR to the equivalent COBOL tokens with the help of pre-defined C-COBOL syntactical token mappings (Table \ref{tab:map}), and (iv) Use Structure Based Traversal (SBT) \cite{Hu2018DeepCC,Ahmad2020ATA} of the IR to generate a sequence of IR tokens(SBT-IR). On similar lines,  we apply transformations (i), (ii) and (iv) to the CodeNet COBOL codes. We further fine-tune UniXCoder \cite{Guo2022UniXcoderUC}, the best-performing model in the literature for zero-shot code-to-code search, with the transformed C codes (C SBT-IRs) from the CodeNet dataset for the code-cloning task. We test this model on the transformed COBOL codes (COBOL SBT-IRs) for clone detection, in a zero-shot setting. 

%krishnam how did you calculate the percentage numbers?
With a best zero-shot test map@2 score of 48.19 and map@1 score of 82.76 for the Code Cloning task with CodeNet COBOL codes, our approach showcased improvement of (a) 32.79 map@2 (212.92\%) and 55.18 map@1 (246.23\%) over a vanilla-transformer \cite{Vaswani2017AttentionIA} auto-encoded with C-Code SBT-IRs for structure learning, and subsequently trained via contrastive loss for learning code semantics, (b) 12.85 map@2 (36.36\%) 
and 24.14 map@1 (45.16\%) over a pre-trained Unix-coder and (c)  11.32 map@2 (30.70\%) and 15.52 map@1 (25.00\%) over a pre-trained Unix-coder fine-tuned on the Code Cloning task with the original C-codes. Following are some of the important observations of our study:   
\begin{enumerate}
    \item (a) demonstrates the efficacy of usage of the model pre-trained on a generic set of PLs, over a model trained from scratch, for better performance on the downstream code-cloning task. 
    \item Though UnixCoder is pre-trained on the code data, it has not seen all the tokens of our IR as well as COBOL tokens. Thus, as depicted in (b), fine-tuning of UniXCoder with C-SBT-IR leads to better performance for the downstream task.
 %   \item (iii) further validates the claim made by UniXCoder of being a better model than CodeT5 for zero-shot transfer.
    \item (c) demonstrates that fine-tuning with C-SBT-IRs helps more for the zero-shot transfer for low-resource COBOL language as compared to fine-tuning with the original C-codes, proving the efficacy of our approach, which uses a common IR across programming languages.
\end{enumerate}

\begin{table}[!t]
\caption{Example C and COBOL codes for a problem description in the CodeNet dataset and corresponding synthesized Intermediate Representations (IR) with Structure Based Traversal (SBT) }
\label{tab:example}
\tiny
\center
\resizebox{\textwidth}{!}{%
\begin{tabular}{|l|lll|}
\hline
\textbf{Problem Description} & \multicolumn{3}{l|}{\begin{tabular}[c]{@{}l@{}}You will turn on the air conditioner if, and only if, the temperature of  the room is 30 degrees Celsius or above. \\ The current temperature of  the room is X degrees Celsius. Will you turn on the air conditioner? \\ Print `'Yes'' if you will turn on the air conditioner;  print `'No'' otherwise.\end{tabular}} \\ \hline
\textbf{C Code} & \multicolumn{1}{l|}{\textbf{Synthesized SBT-IR for C Code}} & \multicolumn{1}{l|}{\textbf{COBOL Code}} & \textbf{Synthesized SBT-IR for COBOL Code} \\ \hline
\begin{tabular}[c]{@{}l@{}}int main()\{\\    int x;\\    \\    scanf("\%d",\&x);\\    \\ \\ \\ \\    \\    if(x\textgreater{}=30)\{\\      \\ \\ \\ \\ \\ \\ \\       printf("Yes");  \\    \\ \\ \\ \\ \\    \}else\{\\       printf("No");  \\    \}\\    \\ \\ \\ \\     return 0;\\ \\ \}\end{tabular} & \multicolumn{1}{l|}{\begin{tabular}[c]{@{}l@{}}(CompUnit(has\_directive(Func\_name(has\_stmt\\     (Block(has\_stmt(Compstmt\\          (has\_stmt (Exprstmt(has\_expr(Call\\              (LI\_name(ACCEPT)ACCEPT)LI\_name\\
(LI\_param(Unary\\                      (Operator(address of)address of)\\                       Operator(U\_expr(Var{[}x{]})Var{[}x{]})U\_expr)\\                              Unary)LI\_param\\          )Call)has\_expr)Exprstmt)has\_stmt\\          (has\_stmt(Ifthen\\               (cond\_expr (Binary\\                   (Operator \\                       (Greater Than Equals)Greater Than Equals\\                   )Operator\\                   (B\_expr1(Var{[}x{]})Var{[}x{]})B\_expr1\\                   (B\_expr2(30)30)B\_expr2\\               )Binary )cond\_expr\\          (then\_stmt (Block(has\_stmt(Compstmt(has\_stmt\\                  (Exprstmt(has\_expr(Call\\                       (LI\_name(DISPLAY)DISPLAY)LI\_name\\                       (LI\_param(Yes)Yes)LI\_param\\                   )Call)has\_expr )Exprstmt\\           )has\_stmt )Compstmt)has\_stmt )Block)then\_stmt\\           (else\_stmt(Block(has\_stmt(Compstmt(has\_stmt\\                (Exprstmt(has\_expr(Call\\                        (LI\_name(DISPLAY)DISPLAY)LI\_name\\                        (LI\_param(No)No)LI\_param\\                )Call)has\_expr)Exprstmt)has\_stmt)Compstmt\\           )has\_stmt)Block)else\_stmt\\           )Ifthen)has\_stmt\\          (has\_stmt(Returnstmt\\               (return\_expr(0)0)return\_expr\\           )Returnstmt)has\_stmt\\     )Compstmt)has\_stmt)Block\\ )has\_stmt)Func\_name)has\_directive)CompUnit\end{tabular}} & \multicolumn{1}{l|}{\begin{tabular}[c]{@{}l@{}}PROCEDURE \\ DIVISION.\\ ACCEPT X.\\ \\ *\textgreater\\        \\ IF (X \textgreater{}= 30)\\        \\ \\ \\ \\ \\ \\ \\ \\ \\        \\ \\       DISPLAY 'Yes'\\ \\ \\ \\ ELSE\\ \\           \\       DISPLAY 'No'\\ \\ \\ \\ END-IF.\\ \\  *\textgreater\\ \\ STOP RUN.\end{tabular}} & \begin{tabular}[c]{@{}l@{}}(CompUnit(has\_directive(Func\_name(has\_stmt\\        (Compstmt(has\_stmt(Exprstmt(has\_expr(Call\\             (LI\_name(ACCEPT)ACCEPT)LI\_name\\             (LI\_param(Var{[}X{]})Var{[}X{]})LI\_param\\         )Call)has\_expr)Exprstmt)has\_stmt\\        (has\_stmt(Ifthen\\            (cond\_expr(Unary\\                 (Operator(()()Operator\\                 (U\_expr(Binary\\                     (Operator\\                     (Greater Than Equals)Greater Than Equals\\                     )Operator\\                     (B\_expr1(Var{[}X{]})Var{[}X{]})B\_expr1\\                     (B\_expr2(30)30)B\_expr2\\                  )Binary)U\_expr\\             )Unary)cond\_expr\\            (then\_stmt(Compstmt(has\_stmt\\                  (Exprstmt(has\_expr(Call\\                         (LI\_name(DISPLAY)DISPLAY)LI\_name\\                         (LI\_param(YES)YES)LI\_param\\                  )Call)has\_expr)Exprstmt\\             )has\_stmt)Compstmt)then\_stmt\\            (else\_stmt(Compstmt(has\_stmt\\                  (Exprstmt(has\_expr(Call\\                         (LI\_name(DISPLAY)DISPLAY)LI\_name\\                         (LI\_param(NO)NO)LI\_param\\                   )Call)has\_expr)Exprstmt\\             )has\_stmt)Compstmt)else\_stmt\\         )Ifthen)has\_stmt\\        (has\_stmt(Exprstmt(has\_expr(Call\\               (LI\_name(exit)exit)LI\_name\\         )Call)has\_expr)Exprstmt)has\_stmt)Compstmt\\ )has\_stmt)Func\_name)has\_directive)CompUnit\end{tabular} \\ \hline
\end{tabular}%
}
\end{table}

\section{Related work}
\textbf{Code Representation Learning:}
Different representations of source code provide distinct perspectives of code understanding. For instance, the Abstract Syntax Tree (AST) provides structural information, Control Flow Graph (CFG) and Data Flow Graph (DFG) provide information about the flow of control and data in the code, and the tokens of the source code itself provide useful syntactic information. Recent studies of neural network-based code learning have tried to learn the semantic latent representation of the code, which are useful for downstream tasks such as code cloning and code-to-code search.  
%Various studies have proposed their own intermediate representations (IR) for source code that are created with the help of source code tokens, AST, CFG, DFG etc. which when learned with neural network models, can be used to accurately understand the functionality of source code and perform various downstream tasks.
CODESCRIBE \cite{guo2022modeling} models the hierarchical syntax structure of code by introducing a triplet position for nodes in the AST. %In this approach, for each node in the AST, its depth, width position of its parent and width position amongst its siblings is recorded and is then incorporated into Transformer and GNN based frameworks. 
GypSum \cite{Wang2022GypSumLH} generates intermediate representation by introducing control-flow related edges %such as Next-Sibling edge, Next-Use-Edge, Child Edge, etc, 
into the AST and then uses graph attention neural networks to generate the encoding. DeepCom \cite{Hu2018DeepCC} converts the input ASTs into specially formatted sequences using a structure-based traversal method. %These sequences are then taken as an input to an attention-based Seq2Seq model. 
% It helps to express the structural information and keep the representations lossless at the same time.
%Both GypSum and DeepCom are trained on a large-scale Java and Python corpus \cite{wan2018improving}. % built from 9,714 open-source projects from GitHub that consists of 87,134 Java code snippets with comments. In addition to this, GypSum is also trained on a Python dataset \cite{wan2018improving} containing 87,226 code snippets	 with comments. 
TPTrans \cite{peng2021integrating} encodes the path between tokens of source code and also the path from leaf to root node for each token in the syntax tree and explores the interaction between them. %They find that a feature overlap between these paths exists and integrate them into a unified transformer framework. It uses the CodeSearchNet dataset \cite{husain2019codesearchnet} for training and evaluation. 
GraphCodeBERT \cite{Guo2021GraphCodeBERTPC} leverages data flow graph for pre-training since it is less complex and hierarchical compared to ASTs.% It uses structure-aware pre-training tasks such as data flow edge prediction and variable alignment for learning representations. 
UniXcoder \cite{Guo2022UniXcoderUC} proposes a one-to-one mapping method to transform AST into a sequence structure while retaining all the structural information of the tree. 
%Krishnam the ASTs etc helps to understand the structure the question is do these models learn the semantics, which can be learnt by contrastive training as what we are doing. That is the loss with which these models are trained? This point is missing. What are the end-tasks of the abovem models? code cloning or code searrch? 
% Krishnam missing literature review should include the following models: %CodeBERT,  GPTC, CBERT, PLBART, ProfetNetCode, CodeGPT, TreeBERT, CodeT5, , Code MVP, SYNCOBERT, DISCO, Corder, CodeRetriever
% Krishnam missing info below
 We notice that all the approaches use huge data for pre-training (in order of $\sim$80 to $\sim$350 GB) as well as fine-tuning for the downstream code-search tasks. However, with the availability of only tiny data for a large legacy language like COBOL, it is impossible to train language models and avail benefits of the above approaches.% On the other hand, our work relies upon language-independent IR to represent code. The model is trained using IRs of a large corpus of C codes and uses the model for the clone detection task of COBOL codes.   
 %Raveendra commented below para
 %However, with the availability of only 727 codes for 325 problem descriptions in CodeNet \cite{Puri2021ProjectCA} dataset, for a legacy language like COBOL, it is impossible to train language models and thus to avail their benefits. In this work we define a meta-model which is instantiated to generate Intermediate Representations (IR) for codes. This meta-model being common across C and Cobol programming languages, we exploit this common representation and test the model trained with the IRs for C codes, for which there is availability of abundant data in CodeNet,  in zero-shot setting for the IRs of COBOL codes for code retrieval task to achieve good zero-shot results.
% End of change 

% Krishnam put missing citations below
% UniXcoder \cite{Guo2022UniXcoderUC} proposes a one-to-one mapping method to transform AST into a sequence structure while retaining all the structural information of the tree. They achieve state-of-the-art performance on code cloning task surpassing ROBERTa \cite{}, CodeBERT \cite{Feng2020CodeBERTAP}, GraphCodeBERT \cite{Guo2021GraphCodeBERTPC}, SynCoBERT \cite{Wang2021SynCoBERTSM}, PLBART \cite{Ahmad2021UnifiedPF}, CodeT5 \cite{Wang2021CodeT5IU}, DISCO \cite{}, Corder \cite{} and CodeRetriever \cite{}. More importantly, UniXcoder achieves state-of-the art results for zero-shot code-to-code search. In this work as we are interested in zero-shot Code Retrieval task for a low-resource COBOL programming language, we choose  UniXcoder as our base model. 

\textbf{Code Clone Detection:}
%Code Clone detection is gaining importance with the increase of source code in software systems. Many Recent works that use pre-trained models for PLs support the task. 
Code Clone detection aims to detect whether two pieces of code have the same semantics or not. Many recent works that use pre-trained models for PLs support the task.  CodeT5 \cite{Wang2021CodeT5IU} learns code semantics through the use of identifier-aware pre-training tasks such as masked identifier prediction (MIP) and identifier-aware denoising, that help the model to distinguish identifiers from other code tokens. PLBART \cite{Ahmad2021UnifiedPF} uses the same architecture as BART along with a denoising autoencoding pre-training task. 
Recently, there have been attempts to improve code semantics through the use of contrastive learning. To generate positive pairs, Contracode \cite{jain2020contrastive} uses transformations such as identifier modifications and code compression (e.g. precomputation of constant expressions) and Corder \cite{bui2021self} uses semantic-preserving transformations such as dead code insertion, statement permutation, identifier renaming etc. To build hard negative pairs, DISCO \cite{ding2022towards}, a self-supervised pre-training model injects real-world security bugs in programs through the misuse of pointers, variables and data-types, and positive pairs are generated by statement permutations and identifier renaming. Syncobert \cite{Wang2021SynCoBERTSM} and CodeMVP \cite{Wang2022CODEMVPLT} build positive pairs using the intermediate representations (IR) of programs such as AST,CFG,DFG which are generated through the compilation process (lexical,syntax,semantic analysis) of the programs. 
% CodeRetriever uses two contrastive learning schemes, unimodal contrastive learning (for code-code pairs) and bimodal contrastive learning (for text-code pairs).
UniXcoder \cite{Guo2022UniXcoderUC} achieves state-of-the-art performance on code cloning task surpassing ROBERTa \cite{liu2019roberta}, CodeBERT \cite{Feng2020CodeBERTAP}, GraphCodeBERT \cite{Guo2021GraphCodeBERTPC}, SynCoBERT \cite{Wang2021SynCoBERTSM}, PLBART \cite{Ahmad2021UnifiedPF}, CodeT5 \cite{Wang2021CodeT5IU}, DISCO \cite{ding2022towards}, Corder \cite{bui2021self} and CodeRetriever \cite{li2022coderetriever}. More importantly, UniXcoder achieves state-of-the-art results for the zero-shot code-to-code search. In this work, as we are interested in a zero-shot Code-Cloning task for a low-resource COBOL programming language, we choose the best-performing UniXcoder as our base model.

\textbf{Cross-language Code Learning:}
Cross-programming language code learning is a relatively unexplored field and there are some recent works that are aimed at learning language-independent representations for source code. \cite{bui2019bilateral} propose a bilateral neural network (Bi-NN) to learn representations of pieces of code in different languages, which can be then used to identify algorithm classes of the code. %They introduce DTBCNN (Dependency Tree based Convolutional Neural Networks) to encode program dependencies as part of AST, which helps to achieve a very high classification accuracy. 
\cite{wang2022unified} propose the Unified Abstract Syntax Tree (UAST) neural network for the cross-language program classification task, by unifying AST traversal and the vocabulary across PLs. %It unifies the AST traversal sequence and its graph-like structure to better capture semantic code features. To achieve the task of cross-language program classification they propose a mechanism called unified vocabulary that reduces the feature gap between different programming languages.
MISIM (Machine inferred code similarity) \cite{ye2020misim} use  a context-aware semantic structure (CASS) %that lifts the semantic meaning from the syntax of code. They use 
and a neural-based code semantics similarity scoring algorithm to learn language-independent representations.
% why the following paper is under cross-lanugage learning
%\cite{bui2021infercode} proposes a self-supervised learning technique for source code called InferCode, which is based on the intuition that similar ASTs have similar subtrees. It uses Tree-based CNNs to generate vectors from ASTs and then uses it to predict the context of subtrees. 
Code Transformer \cite{zugner2021language} learns language-agnostic features from both, the structure and the context of programs %. Due to the language-agnostic nature of the model,
by training with multiple PLs, which leads to larger improvements in performance for languages with lower resources. However, these approaches which  learn unified representations across programming languages require some amount of data for each language and thus prove not to be useful for legacy languages like COBOL in a zero-shot setting. %On the other hand, our paper introduces a novel Intermediate Representation that is common to both Cobol and C programming languages. Having a common IR significantly enhances cross-language representation learning and thus leads to good performance on zero-shot code-retrieval task with COBOL.

\section{Dataset}
We form our dataset using the C and COBOL codes available in CodeNet \cite{Puri2021ProjectCA}. The dataset consists of 4053 Problem Descriptions (PD) (as a .html file) that have a combined total of 313360 C codes. The codes which belong to the same problem description form clones of each other as they share the same semantics. CodeNet is pre-processed by removing the PDs with: (i)  no or empty .html files (ii) no accepted (correctly executing to provide the right output) C codes (771 PDs) (iii) only one C code (to detect clones there must be at the least two C codes per PD to form a positive pair). Thus, the final dataset consists of 3221 PDs with a total of 303193 C codes.% wherein each PD contains at least two codes. 

Generating the IRs and SBTs for the codes (explained in section \ref{sec:app}) using our meta-model is a time-consuming task. It takes on an average of $\sim$3 seconds per code. % to generate the IR and then the Structure Based Traversal (SBT) representation for the same. We run our IR and SBT generation algorithms explained in the next section for 
We generate IRs and SBTs of C codes belonging to randomly selected 1693 PDs ($\sim$50\% of 3221 PDs - taking us 10 days) and create the train-val-test splits (Table \ref{tbl:dataset}). % Thus, we have SBT-IR generated for C codes belonging to 1693 PDs. 
Though we are using partial data, using all the C-codes available in CodeNet  dataset would only lead to improvement in our overall results. % We use these C-SBT-IRs belonging to 1693 PDs to create the train-val-test splits. 
The token length of generated SBT-IRs is much more than the original C-codes. This is because, in addition to the leaf-nodes of the AST based IR, which are the actual code tokens, the SBT-IRs have the non-leaf tokens as well. Since the maximum sequence length after tokenization accepted by UniXcoder model is 512; we create a separate dataset of codes whose SBT-IRs  fit into this token length. This allows us to evaluate the drop in the performance of our approach due to truncated SBT-IRs for the codes. %Table \ref{tbl:dataset} denotes the splits created from the CodeNet dataset using the SBT-IRs of 1693 PDs.
%and following sections explain the process we followed for creating the splits. 

We use MAP@R as the metric to evaluate the performance of the Code-cloning task. Each problem statement in the test set consists of a total of `R + 1' codes which have the same semantics. For each code, `R' most semantically  similar codes are retrieved from the test set based on the model's predictions and the average precision score is evaluated. The mean of all the average precision scores is then taken to get the final MAP@R score. We form our test splits such that we can compute MAP@R for certain value of R based on the availability of data. 

\subsection{Creating Test Dataset for COBOL Code Cloning}
 %Out of 727 accepted COBOL submissions for 325 problem descriptions, there are 92 PDs which have 3 accepted COBOL codes each. We use these 276 COBOL programs as our COBOL zero-shot test set, which allows us to compute MAP@2 per PDs (The final score is averaged over all PDs). We further remove these 92 PDs from the above 1698 PDs for which at the least two C codes are available, to avoid information leakage, resulting in 1606 PDs. We split these in approximately 90:10 split resulting in 1436 PDs ($\sim$200K C Codes) in train and 170 PDs ($\sim$18K C Codes) validation split. We also perform an experiment with codes fitting in 512 token length which is maximum allowable token length by UniXCoder and CodeT5 model. This data contains $\sim$22K C codes belonging to 149 PDs in training, $\sim$1K C codes belonging to 20 PDs in validation and 58 COBOL codes belonging to 29 PDs in zero shot test-split (Here we compute MAP@1).

%We use MAP@R as the metric to evaluate the performance of the Code-cloning task. Each problem statement in the dataset consists of a total of `R + 1' codes which have the same semantics. For each code, `R' most semantically  similar codes are retrieved from the dataset based on the model's predictions and the average precision score is evaluated. The mean of all the average precision scores is then taken to get the final MAP@R score.
%Krishnam  provide the definition / formula and explaination of MAP@K in detail
%We form our test splits such that we can compute MAP@R for certain value of R based on the availability of data. 

 Out of 727 accepted COBOL submissions for 325 PDs in CodeNet there are there are 92 PDs that have 3 or more accepted COBOL codes.  We randomly sample 3 codes from each one of these 92 PDs to obtain a total of 276 COBOL codes forming our zero-shot COBOL test set. As each of the PDs in this test set has 3 COBOL codes, this allows us to compute MAP@2 for this test set. Henceforth, we refer to this test set as \textit{COBOL-Test-MAP@2}. For generating the second test set of COBOL codes that fit into  512 tokens length, we filter out the SBT-IRs of the 727 accepted COBOL codes that exceed this limit. There are 29 PDs for which at least 2 accepted COBOL codes are present that do not exceed this limit of 512 token length. We sample 2 codes randomly from each of these PDs to obtain a total of 58 codes. This allows us to compute MAP@1 for this test set. Henceforth, we term this test split of COBOL as \textit{COBOL-Test-MAP@1}.

\begin{table}[]
\caption{Created Train-Test Splits from CodeNet Dataset \cite{Puri2021ProjectCA}}
\begin{center}
\scalebox{0.75}{
 \begin{tabular}{ p{4cm}|p{2cm}|p{1.5cm}|p{1.5cm}|p{2cm}}
 \hline
Splits    & Token Length & PDs &Codes & Codes per PD\\
 \hline
 %  \multicolumn{6}{c}{C Codes} \\
 % \hline
 \textit{Train-C-ALL}  & ALL & 1436 & 199675 & Average 139 \\
  \textit{Val-C-ALL}  & ALL & 170 & 18186 & Average 107\\
  \textit{Test-C-MAP@299} & ALL & 29 & 8700 & 300 \\
    \textit{Test-COBOL-MAP@2} & ALL & 92 & 276 & 3\\
  %C&All & All & 1635 & 226516 & - \\
 \hline
  \textit{Train-C-512}& 512 & 201 & 33693 & Average 167\\
  \textit{Val-C-512}  & 512 & 30 &  3066 & Average 102\\
\textit{Test-C-MAP@99} & 512 & 11 & 1100 & 100 \\
  %C&All & 512 &  &  & - \\
  \textit{Test-COBOL-MAP@1} & 512 & 29 & 58 & 2\\
 \hline
 %  \multicolumn{5}{c}{COBOL} \\
 % \hline

% COBOL & Test & ALL & 92 & 276 & 3\\

% COBOL & Test & ALL & 92 & 276 & 3\\
% \hline

\end{tabular}} \label{tbl:dataset}
\end{center}
\vspace{-2mm}
\end{table}

\subsection{Creating Dataset for C Code Cloning}
% We randomly split the 1698 PDs in 80:10:10 split to get $\sim$200K C Codes in train (1360 PDs), $\sim$30K codes in validation (180PDs) and remaining codes in the test split. From the PDs of the codes in the test split we choose 29 PDs each having at the least 300 accepted C codes. We randomly select 300 C codes for each of these 29 PDs (total 8700 codes), which allows us to compute MAP@299 for this test split. For the experiment with codes fitting in 512 maximum token length by the models we have $\sim$34K C codes belonging to 201 PDs in train, $\sim$3K C codes belonging to 30 PDs in validation and 1100 codes belonging to 11 PDs in test split (Here we compute MAP@99).

 To create the train-val-test splits for C code IR-SBTs for 1693 PDs (discussed prior in this section), we first remove all the C codes belonging to the PDs used to form the COBOL test sets. This is done to ensure that there is no information leakage in the train-val splits of C with the test splits of COBOL in the zero-shot test setting. Taking the union of the PDs from the two COBOL test splits, we have 99 PDs. There are 87 PDs which are common in 1693 PDs for which we have C SBT-IRs and the 99 COBOL test PDs.  We split the remaining 1606 PDs in the ratio of 90-10 to form the train-validation splits for the C codes, which we refer to as \textit{Train-C-ALL} and \textit{Val-C-ALL}, respectively.  We use the remaining 87 PDS to form the test split for C.  We observe that 29 out of 87 PDs, have at least 300 C codes available. We form the test split of C (\textit{Test-C-MAP@299}) by randomly sampling 300 codes from these 29 PDs.  For the second test split for C,  we choose 11 PDs out of 87 PDs that have at least 100 C codes each with a maximum token length of 512. We term this split as \textit{Test-C-MAP@99}. The details of the splits are provided in Table~\ref{tbl:dataset}.

\begin{figure} [t!]
 \centering
 \includegraphics[width=\linewidth]{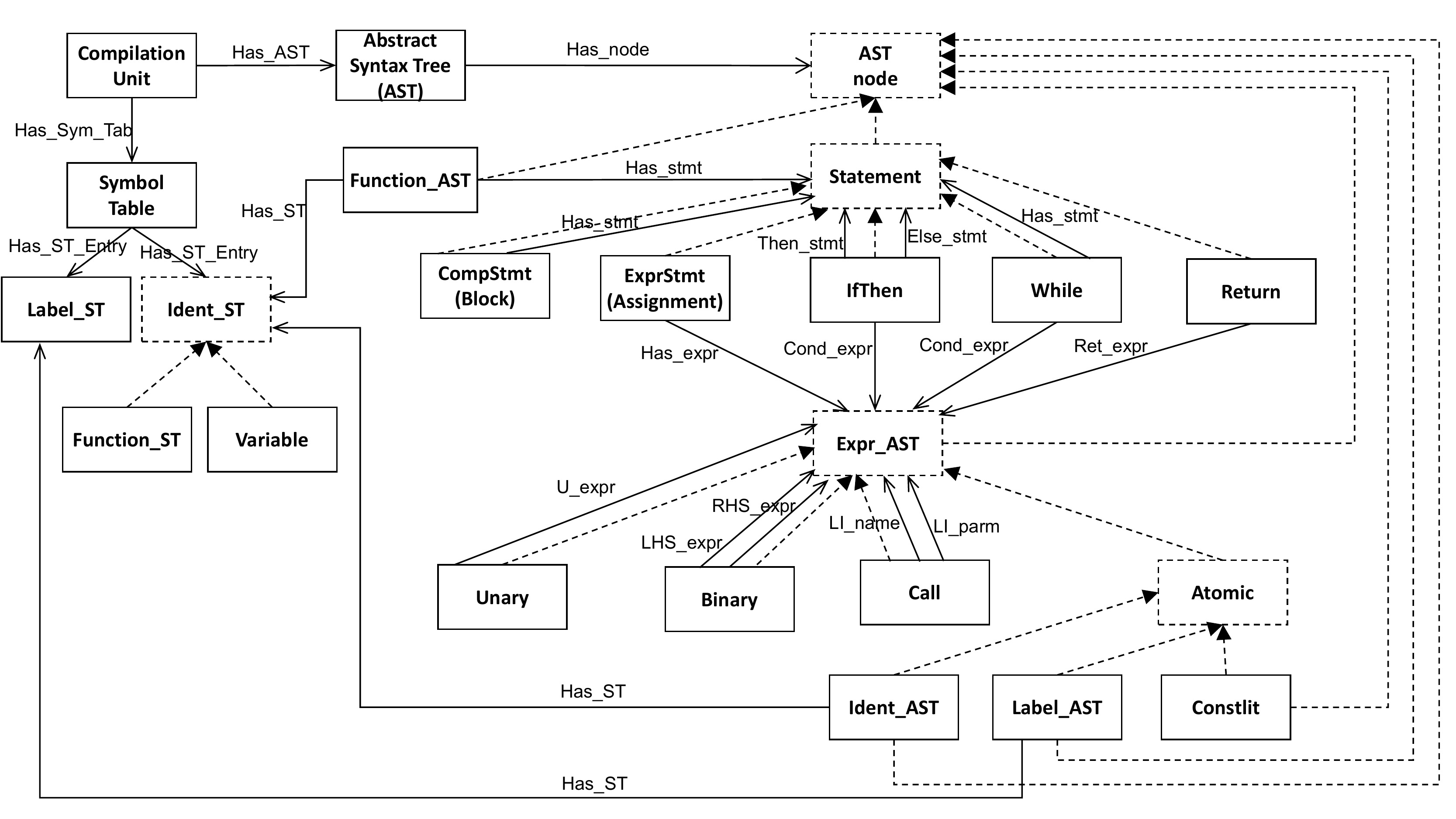}
 \caption{IR Meta model}
 %\Description{Figure description}
 \label{fig:meta-model}
 \vspace{-2mm}
\end{figure}

\begin{figure} [t!]
 \includegraphics[width=\linewidth]{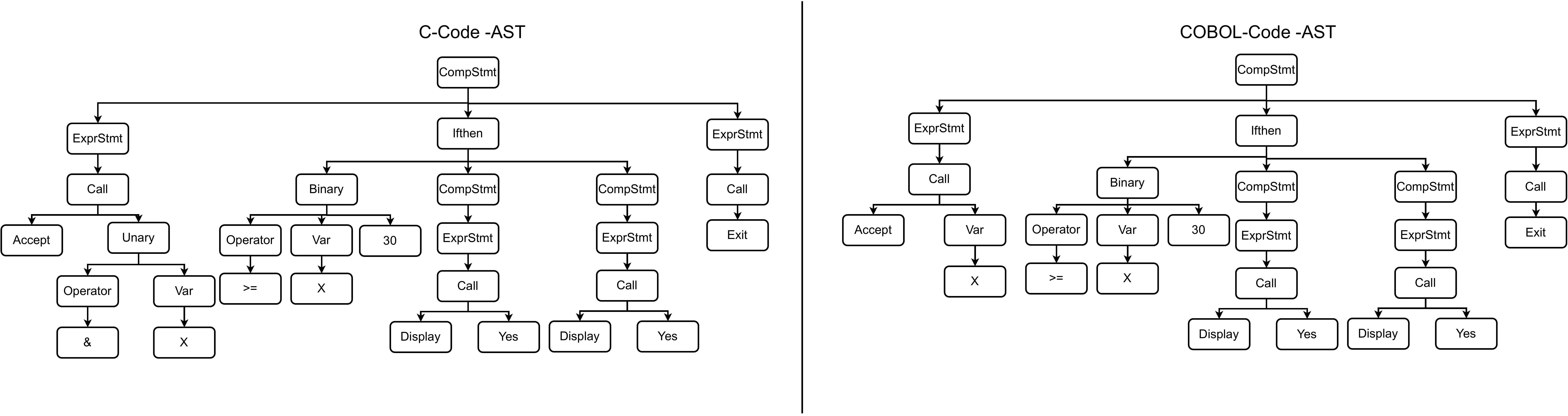}
 \caption{ASTs for the C and COBOL codes in Table 1}
 %\Description{Figure description}
 \label{fig:AST}
 \vspace{-2mm}
\end{figure}

\section{Approach}\label{sec:app}
{\textbf{Intermediate Representation}}: Figure~\ref{fig:meta-model} illustrates key parts of our IR meta-model, which is common for both C and COBOL PLs. The meta-model is designed in such a way that language-specific C and COBOL syntactic constructs having the same semantics are represented using similar IR elements (\emph{objects}). This design is influenced by static program analysis tasks. %for which it was originally designed. 
At a high level, our meta-model captures different kinds of IR elements as \emph{classes}(represented in boxes in the Figure) and relationships among them (represented as arrows). In this Figure, the dotted boxes represent abstract classes and the solid boxes represent concrete classes. The dotted arrows represent the generalization relationship between abstract and concrete classes, whereas solid arrows represent various relationships among the concrete classes. An IR of a program is constructed by instantiating relevant IR elements in the meta-model. A program in a file is represented as \emph{compilation unit} class that consists of an \emph{Abstract syntax tree}(AST) and a \emph{Symbol table}(ST). AST is a tree of \emph{nodes} (ASTNode), where each node represents an abstract class in the program. The symbol table stores symbols used in the program. There are several \emph{types} of nodes such as statements, expressions, functions, variables used in expressions, labels, and constants used in the program. The central elements are \emph{statements} and \emph{expressions}. They are further classified into different types and the relationship among them is shown in Figure~\ref{fig:meta-model}. For example, consider a statement $x = y\;$. It has a \emph{ExprStmt} node representing the whole statement, a \emph{Binary} node with `=' as its operator, with \emph{Ident\_AST} node with node symbol \emph{x} as its \emph{LHS\_expr} and another \emph{Ident\_AST} node with symbol \emph{y} as its \emph{RHS\_expr}. The symbols (variables, labels, and constants) used in the expressions are linked to corresponding symbol table objects. For example, the same variable \emph{x} used in two different expressions has two \emph{Ident\_AST} nodes in AST but linked to one symbol \emph{x} in the symbol table. 

The C and COBOL language front-ends parse the program and transform the concrete syntax tree (CST) into IR by instantiating relevant objects from their classes and relationships of the meta-model. For our work, we only consider the AST of the IR. For semantically equivalent COBOL and C codes in Table \ref{tab:example} their corresponding ASTs are illustrated in Figure~\ref{fig:AST}. We further linearize the generated AST using the Structure-Based Traversal (SBT) of the AST. The SBT algorithm starts with the root node of the AST and recursively traverses the child nodes using Depth First Search (DFS). The algorithm emits the node types at the entry and exit of a visiting node. If the node is an intermediate node of a tree, it proceeds to process the child node. If the node is a leaf node, the algorithm emits the leaf value after applying the transformations given in Table~\ref{tab:map}, if required. The algorithm terminates when all the nodes of the AST are visited. Examples of sequences generated by SBT traversal of a COBOL and a C code IRs (SBT-IRs) are illustrated in Table~\ref{tab:example}.

\begin{table}[t]
\caption{C-COBOL Token Mapping}
\label{tab:map}
\begin{center}
\scalebox{0.75}{
\begin{tabular}{ll|ll}
%\multicolumn{1}{c}{\bf C Code Tokens}  &\multicolumn{1}{c}{\bf COBOL Code Tokens}&\multicolumn{1}{c}{\bf C Code Tokens}&\multicolumn{1}{c}{\bf COBOL Code Tokens}\\ \hline \\
\bf C Code Tokens  & \bf COBOL Code Tokens & \bf C Code Tokens &\bf COBOL Code Tokens
\\\hline 
scanf  &ACCEPT & printf & DISPLAY\\
strtok & UNSTRING &  , & DELIMITED\\
= & INTO &  strlen & LENGTH OF\\
strcat& STRING &  strlen & STORED-CHAR-LENGTH\\
qsort & SORT & strlen & COUNT\\
fread & READ & stdin/stdout & CONSOLE\\
lsearch/ bsearch & SEARCH & statistical & ORD\\
\% & REM/ MOD & round & ROUNDED\\
+ & SUM & memset & INITIALIZE\\
\end{tabular}}
\end{center}
\vspace{-2mm}
\end{table}
While transforming the C Codes, we replace the C code tokens listed in Table \ref{tab:map}, which appear as the leaf nodes of the AST (IR), with the semantically equivalent COBOL tokens to facilitate zero-shot transfer. Depending on language syntax, distinct surface forms of the codes written for the same PD may use distinct functions and variable names. To retain consistency across distinct surface forms and to avoid the inductive bias created by semantically meaningful function and variable names in both C and COBOL codes, we replace them with generic tokens such as $Func$ and $VAR$ (Refer to Table \ref{tab:example} SBT-IR for an example). We ensure that this replacement does not hamper the data and control flow of the code, by keeping track of original names and replacing every instance of it with a distinct generic names for each of the original names.  This transformation forces the model to learn the semantic similarity between positive code pairs by understanding the underlying logic of the code and not by just exploiting the mappings between the semantically meaningful function and variable names to learn the similarity. 

To fine-tune the UniXCoder \cite{Guo2022UniXcoderUC} model for Code-Clone detection, we form the positive and negative pairs of the C codes in \textit{Train-C-ALL} and \textit{Val-C-ALL}. The positive (clones) and negative (not-clones) pairs are the codes belonging to the same and distinct problem descriptions, respectively. The task is treated as a binary classification task, where the class label is 1 if the pair of codes provided are clones of each other or is 0, otherwise. The UniXCoder \cite{Guo2022UniXcoderUC} model is trained in an encoder-only setting, with the cross-entropy loss, a batch size of 8, and a learning rate of 5e-5. We save the model with the best F1-score on the validation set \textit{Val-C-ALL}. We repeat the experiments with (i) original C-codes  (ii) C-SBT-IRs  and (iii) C-SBT-IRs truncated with maximum allowable token length by UniXcoder (512)  (\textit{Train-C-512} and \textit{Val-C-512}). 

We also train a vanilla transformer model in an auto-encoder setting by reconstructing the input C-SBT-IRs with a reconstruction loss for making the model learn the syntax and then fine-tune the trained encoder of the transformer in a Siamese setting with a contrastive loss using positive and negative pairs of code SBT-IRs as explained above to make the model understand the code semantics. Thus, we have a total of four models trained: (i) \textit{UniXCoder-C-SBT-IR-ALL}: UniXcoder fine-tuned with all C-SBT-IRs (ii) \textit{UniXCoder-C-SBT-IR-512}: UniXcoder fine-tuned with C-SBT-IRs which fit into 512 token length (iii) \textit{Transformer-C-SBT-IR-ALL}: Vanilla transformer  trained with all C-SBT-IRs (iv) \textit{UniXCoder-C-Code-ALL}: UniXcoder fine-tuned with all C codes. We perform inference with our four test splits described in the prior section: (i) \textit{Test-COBOL-MAP@2}, (ii) \textit{Test-COBOL-MAP@1}, (iii) \textit{Test-C-MAP@299}, (iv) \textit{Test-C-MAP@99}.

\begin{table}[]
\caption{Results}
\begin{center}
\scalebox{0.75}{
 \begin{tabular}{ p{4cm}||p{3.2cm}|p{1cm}||p{3.2cm}|p{1cm}}
 \hline
Trained Model   & Test Split &   MAP & Test Split   &   MAP\\
 \hline
  Random & \textit{Test-COBOL-MAP@2}  & 0.54  & \textit{Test-COBOL-MAP@1} &  1.72\\
\textit{Transformer-C-SBT-IR-ALL}&\textit{Test-COBOL-MAP@2}  & 15.40 &\textit{Test-COBOL-MAP@1}&22.41\\
 \textit{UniXCoder-Pretrained} &\textit{Test-COBOL-MAP@2} & 35.34&\textit{Test-COBOL-MAP@1} &  53.45\\
\textit{UniXCoder-C-Code-ALL} & \textit{Test-COBOL-MAP@2}&  36.87 &\textit{Test-COBOL-MAP@1}& 62.07\\
 \textit{UniXCoder-C-SBT-IR-ALL}&\textit{Test-COBOL-MAP@2} &  \textbf{48.19} &\textit{Test-COBOL-MAP@1}& 77.59\\
 \textit{UniXCoder-C-SBT-IR-512} &\textit{Test-COBOL-MAP@2} & 45.56 &\textit{Test-COBOL-MAP@1}& \textbf{82.76}\\
 \hline
  Random & \textit{Test-C-MAP@299}  & 0.19 & \textit{Test-C-MAP@99} &  1.23\\
\textit{Transformer-C-SBT-IR-ALL}& \textit{Test-C-MAP@299} & 31.05 &\textit{Test-C-MAP@99} & 40.55\\
 \textit{UniXCoder-Pretrained} & \textit{Test-C-MAP@299}& 27.63& \textit{Test-C-MAP@99} &  46.78\\
\textit{UniXCoder-C-Code-ALL} &\textit{Test-C-MAP@299} &  32.91 &\textit{Test-C-MAP@99} & 52.94\\
 \textit{UniXCoder-C-SBT-IR-ALL}& \textit{Test-C-MAP@299}&  \textbf{64.75} &\textit{Test-C-MAP@99} & 89.12\\
 \textit{UniXCoder-C-SBT-IR-512} & \textit{Test-C-MAP@299}& 63.44 &\textit{Test-C-MAP@99} & \textbf{90.82}\\
 \hline
\end{tabular}} \label{tbl:result}
\end{center}
\vspace{-4mm}
\end{table}

\section{Results and Discussions}\label{sec:rad}
%Krishnam you need to explain how we calculated the random score 
% some observations below elaborate on the same
Table \ref{tbl:result} shows the performance comparison on all the test datasets. We use the algorithm by \cite{mou2016convolutional} for MAP@R score computations. 
%Krishnam provide the formula of MAP score computation here
Testing for COBOL is done in a zero-shot setting as training is performed only by using C-SBT-IRs or C codes. We have computed random MAP scores as one of the benchmarks. Random MAP is the MAP score that is obtained if the selection of similar codes in the test split is done in a random manner. 
% Krishnam the sentence below - I am not understanding
%Randomized predictions (of code pairs) are generated. 
 For both the COBOL test splits, the average is taken for ten-thousand runs to get the final scores as 0.54 for \emph{Test-COBOL-MAP@2} and 1.72 for \emph{Test-COBOL-MAP@1}. The C test splits being larger than COBOL, the average is taken over thousand runs to get the final scores as 0.19 for \emph{Test-C-MAP@299} and 1.23 for \emph{Test-C-MAP@99}. As illustrated in Table \ref{tbl:result}, our results are significantly better than random MAPs on all test splits.

% This score is inversely proportional to the number of problem statements in the dataset i.e. the larger the number of problem statements, the lower is the random MAP score for that dataset.
% It is also a measure of the difficulty to get a higher score by the model. The lower the random MAP score, the more difficult it is for the model to perform better.

For both test splits, the performance of the pre-trained UniXCoder is significantly better than that of the vanilla transformer trained using C-SBT-IRs. The MAP scores for \emph{Test-COBOL-MAP@2} and  \emph{Test-COBOL-MAP@1} are increased by the magnitude of 19.94 and 31.04, respectively. This demonstrates the efficacy of the use of a model pre-trained with a large amount of code data yielding superior performance over a model trained from scratch with a comparatively small amount of data.  
%This indicates that UniXCoder is able to learn the SBT representations and performs much better on understanding tasks like code clone detection. 

There is a very small amount of improvement in the performance (1.53 for \emph{Test-COBOL-MAP@2} and 8.62 for \emph{Test-COBOL-MAP@1}) after task-specific fine-tuning of UniXCoder with C-codes over the pre-trained model. This shows that the model fine-tuned with C-codes is not generalizable for unseen COBOL codes.  %there is not much cross-language learning while training with the source codes.
UniXCoder trained on C-SBT-IRs yields the best results on the COBOL test splits in the zero-shot setting. There is an increase in MAP score of 12.85 (36.36\% rise) for \emph{Test-COBOL-MAP@2}, 24.14 (45.16\% rise) for \emph{Test-COBOL-MAP@1} % 37.12 (134.35\% rise) for \emph{Test-C-MAP@299} and 44.04 (94.14\% rise) for \emph{Test-C-MAP@99} 
over the pre-trained UniXCoder without any programming language-specific fine-tuning. This demonstrates that the model trained with the common SBT-IR representations of C code facilitates the transferability of code understanding from C to COBOL, yielding much better zero-shot performance on COBOL.  %Since the model accepts a maximum sequence length of 512 and truncates the sequence beyond this limit, we have created two train sets: \emph{Train-C-ALL} and \emph{Train-C-512} from table 2. 
The drop in performance on \emph{Test-COBOL-MAP@2}, when tested on the model trained with  \emph{Train-C-SBT-IR-512} (45.56\%) as opposed to the model trained with the complete data (\emph{Train-C-SBT-IR-ALL}) (48.56\%), is very less. All the above results showcase that the C-SBT-IRs fitting into maximum token lengths provides better supervision than the ones which don't. This can be because of the ones which exceed the maximum token length, on truncation, might be losing the semantic information and thus may not provide correct supervision for the code-cloning task. Moreover, the results show that the model 
trained with \emph{Train-C-ALL} yields the best results for \emph{Test-COBOL-MAP@2} that is the test set with no restriction on the token lengths. Similarly, when the model is trained with C-IR-SBTs that have lesser than 512 sequence length i.e. \emph{Train-C-SBT-IR-512} it gives the best results for \emph{Test-COBOL-MAP@1} that is the test set containing COBOL-IR-SBTs with sequence length lesser than 512.  Thus, there is a huge performance gap (MAP@2 of 48.19 vs MAP@1 of 82.76) between the results of the COBOL codes not fitting into 512 token lengths, versus the COBOL codes fitting into the maximum token length. This performance gap can be attributed to the fact that truncation of the test COBOL codes exceeding the sequence length limit will lead to lesser understanding of the code semantics by the model and also a higher random MAP value for the MAP@1 test split.
%Krishnam editing here
% Again, this can be because of the truncation on the test COBOL codes exceeding the limit leading to less understanding of the code semantics by the model.

Similar observations can be made on the test splits of C-SBT-IRs. Moreover, UniXCoder fine-tuned with C-SBT-IRs gives better results for C-SBT-IRs as compared to COBOL-SBT-IRs. This slight deprecation in overall scores for COBOL over C is due to the inherent difference in the structure of code in the two programming languages, leading to a model trained with C-SBT-IRs performing better on C-SBT-IRs as compared to COBOL-SBT-IRs. %The meta-model maps similar constructs to each other from the two languages but it does not mean that there is an exact relationship between them.
The Random MAP scores for C test splits are also lower than that for COBOL which makes the results on C even stronger.

\section{Conclusion}
We define a neuro-symbolic approach to address the Code-Cloning task for legacy COBOL codes. We define a meta-model that is instantiated to generate an Intermediate Representation common across no-resource COBOL and high-resource C codes. We use the C code IRs to fine-tune the UniXCoder model for the Code-Cloning task and perform inference on IRs of COBOL codes. The fine-tuned model leads us to a performance improvement of 12.85 MAP@2 for COBOL over the pre-trained UniXCoder model in zero-shot setting and an improvement of 32.69 MAP@2 over a vanilla transformer model trained with C code IRs. This demonstrates the efficacy of our approach of performing zero-shot transfer with the common IR and also the use of the model pre-trained on a lot of code data over a model trained from scratch to better facilitate the transfer. 

% Lovekesh - we can add the para below instead of defending about not using LLMs in the intro
\section{Furture Work}
More recently, Large Language Models (LLMs) such as T5\cite{48643}, GPT3\cite{brown2020language}, Codex\cite{Chen2021EvaluatingLL}, PaLM\cite{chowdhery2022palm}, pre-trained with massive volumes (in the order on 50TB) of data have shown improved performance for code understanding tasks using in-context learning \cite{Brown2020LanguageMA,Huang2022TowardsRI}, which work in zero-shot or few-shot setting, with no or very less number of available samples for a given task. However, in-context learning does not work well with smaller Language Models (LMs) \cite{Qiu2022EvaluatingTI,Hosseini2022OnTC} and they require task-specific fine-tuning to yield acceptable performance. %Thus, LLMs can provide  zero or few-shot solution to perform various tasks for low-resource languages.  %However, eventually these LLMs are not going to be available as cost-free solutions and also would require porting of the propriety code data to the server on which these LLMs reside, raising the recurrent cost and code privacy concerns. On the other hand, above listed smaller pretrained LMs are freely available and can reside on local server. 
As future work, we want to utilize the in-context learning capabilities of LLMs, which is the upcoming paradigm to address code understanding in low-resource setting. %In this work, meta-model based intermediate representation, allowed us to utilize the transformed C-IR data to fine-tune UnixCoder LM and transfer learn in zero-shot setting for COBOL-IRs. As the future work,
We plan to compare our UniXCoder based zero-shot performance of COBOL code-cloning task facilitated by a common meta-model and IR, with a zero-shot performance for the task using LLMs pre-trained with a lot of code data.

\bibliographystyle{unsrtnat}
\bibliography{references}  %%% Uncomment this line and comment out the ``thebibliography'' section below to use the external .bib file (using bibtex) .

\end{document}